\documentclass{article} 
\usepackage{onecol}

\usepackage{times}
\usepackage{microtype}
\usepackage{epsfig}
\usepackage{graphicx}
\usepackage{amsmath}
\usepackage{amssymb}
\usepackage{algorithm}
\usepackage{algorithmic}

\usepackage{color}
\usepackage{ccicons}    
\usepackage{txfonts}
\usepackage{textcomp}

\usepackage{url}
\usepackage{multirow}
\usepackage{booktabs}
\usepackage{mathtools}
\usepackage{tabularx}
\usepackage{varwidth}
\usepackage{stmaryrd}
\usepackage{balance}

\def\argmin{\mathop{\rm arg\,min}\limits}%    a math operator.

\makeatletter
\DeclareRobustCommand\onedot{\futurelet\@let@token\@onedot}
\def\@onedot{\ifx\@let@token.\else.\null\fi\xspace}

\def\etal{\emph{et al}\onedot}
\makeatother

\usepackage[pagebackref=true,breaklinks=true,letterpaper=true,colorlinks,bookmarks=false]{hyperref}
\begin{document}

%%%%%%%%% TITLE
\title{Ranking CGANs: Subjective Control over Semantic Image Attributes}
%\author{Yassir~Saquil$^{1}$, Kwang~In~Kim$^{1}$ and Peter~Hall$^{1}$\\
%$^1$University of Bath \\
%}
\author{Yassir~Saquil, Kwang~In~Kim, and Peter~Hall\\
University of Bath \\
}
\date{}

\maketitle

\begin{abstract}\noindent
In this paper, we investigate the use of generative adversarial networks in the task of image generation according to subjective measures of semantic attributes. Unlike the standard (CGAN) that generates images from discrete categorical labels, our architecture handles both continuous and discrete scales. Given pairwise comparisons of images, our model, called RankCGAN, performs two tasks: it learns to rank images using a subjective measure; and it learns a generative model that can be controlled by that measure. RankCGAN associates each subjective measure of interest to a distinct dimension of some latent space.
We perform experiments on UT-Zap50K, PubFig and OSR datasets and demonstrate that the model is expressive and diverse enough to conduct two-attribute exploration and image editing.
\end{abstract}

%%%%%%%%% BODY TEXT
\section{Introduction}

Humans routinely subjectively  order objects on the basis of semantic attributes. For example,  people agree on what is ``sporty'' or ``stylish'', at least to a sufficient extent that meaningful communication is possible. Hence one can imagine applications such as  a ``shopping assistant'' in which a user can browse a collection using subjective terms ``show me something more elegant than the dress I am looking at now".
However, such subjective concepts are ill defined in a mathematical sense, making the building of a computational model of them a significant open challenge.

In this paper, we propose a neural architecture that addresses the problem of synthesizing natural images by controlling subjective measures of semantic attributes. For instance, we want a system that produces rank ordered images of shoes according to the subjective degree to which they exhibit a semantic attribute such as ``sporty", or pairs of such attributes --  ``comfortable'' and ``light'', for example.

The underlying mechanism is a Conditional Generative Adversarial Network (CGAN) \cite{CGAN}. CGANs provide a low dimensional latent space for realistic image generation where the images are synthesised by sampling from the latent space. However, controlling CGAN to generate an image with a specific feature is challenging, because the input to CGAN is an $n-$dimensional noise vector. Our work shows it is possible to decompose this latent space into subspaces: one is a random space as usual in CGAN, the other subspace is occupied by control variables -- in particular subjectively meaningful control variables.

In recent work, semantic (subjective) attributes were defined as categorical labels \cite{CatAtt,CatAtt2} indicating the presence or the absence of some attributes. A conditional generative model can be trained, under supervision, where the labels are assigned to a latent variable. This results in some control over the generated images, but that control is limited to switching on or off the desired attribute: it is binary. In contrast, we provide a mapping of semantic attributes onto a continuous subjective scale.

Training is a particular issue for ranked subjective attributes. Although humans can order elements in an object class, there is typically no scale associated with the ordering: people can say one pair of shoes is more ``sporty'' than another pair, but not by how much. To address this problem we train using an annotated set of image pairs where each pair is ordered under human supervision. A ranking function is learned per attribute which can predict the rank of a novel image.

The key characteristic of our approach is the  addition of a {\em ranking} unit that operates alongside the usual discriminator unit. Both the ranker and the discriminator receive inputs from a generator. As is usual, the role of the discriminator is to distinguish between real and fake images; the role of the ranker is to infer subjective ranking according to the semantic attributes. We call our architecture RankCGAN.

We evaluated RankCGAN on three datasets, shoes (UT-Zap50K) \cite{Zappos}, faces (PubFig) \cite{Pubfig} and scenes (OSR) \cite{Relative} datasets. %Results show that our model can represent a continuous variation of attributes and disentangle between them. Also, it keeps a correct variation of the attribute's strength with respect to a ranking score which is not guaranteed with a standard CGAN.
Results in Section~\ref{sec:results} show that our model can disentangle multiple attributes and can keep a correct continuous variation of the attribute's strength with respect to a ranking score, which is not guaranteed with a standard CGAN.

\textbf{Contributions}: 
In sum, our contributions are: (1) To provide a solution to the problem of the subjective rank ordering of semantic attributes; (2)  A novel conditional generative model called RankCGAN, that can generate images under semantic attributes that are subject to a global subjective ranking; 
(3) A training scheme that requires only pairs of images to be subjectively ranked. There is no need for global ranking or to annotate the whole dataset.

The experimental section, Section~\ref{sec:results}, provides evidence for these claims. We show quantitative and qualitative results, as well as applications in  attribute-based image generation, editing and transfer tasks.

%------------------------------------------------------------------------
\section{Related work}
\paragraph{Deep Generative Model:}
There is substantial literature on learning with deep generative models. Early studies were based on unsupervised learning by using restricted Botlzmann machines and denoising auto-encoders \cite{Hinton2,Hinton1,Vincent,Srivastava}. Recently, deterministic networks \cite{Chairs,DET1,DET2} propose architectures for image synthesis. In comparison, stochastic networks rely on a probabilistic formulation of the problem. The Variational Autoencoder (VAE) \cite{VAE,rezende} maximizes a lower bound on the log-likelihood of training data. Autoregressive models {\em i.e.} PixelRNN \cite{pixelRNN} represents directly the conditional distribution over the pixel space. Generative Adversarial networks (GANs) \cite{GAN} have the ability to generate sharp image samples on datasets with higher resolution.

Several studies have investigated conditional image generation settings. Most of the methods use a supervised approach such as text, attributes and class label conditioning to learn a desired transformation \cite{CGAN,CPixelRNN,Ccaption,Text2Image}. Additionally, there are works on image-conditioned models, such as style transfer \cite{Style1,Style2}, super-resolution \cite{SuperRes1,SuperRes2} and cross-domain image translation \cite{Pix2pix,AlignGAN,Cross3,Cross4}.

In contrast, fewer works have focused on decomposing the generative control variable into meaningful components. InfoGAN \cite{InfoGAN} uses an unsupervised approach to learn semantic features  maximizing mutual information between the latent code and the generated observation. In the spectrum of supervised approaches, VAE \cite{VAE} was used in DC-IGNs \cite{DC-IGN} to learn latent codes representing the rendering process of 3D objects, similarly used in Attribute2Image \cite{Attribute2Image} to generate images by separately generating the background and the foreground. Attempts to study the incorporation of the adversarial training objectives were conducted in VAE \cite{GAN+VAE} and autoencoders \cite{AAE} settings. The fader network \cite{Fader} approach is an end-to-end architecture where the attribute is incorporated in the encoded image. Trained using categorical labels, it controls the presence of an attribute while preserving the naturalness of the image, which requires a careful parametrization during the training process since the decoder is updated on the reconstruction and adversarial objectives.
Independently, CFGAN \cite{CFGAN} proposes an extension of CGAN, where the attribute value is not fed directly to the model but associated to latent vector using a filtering architecture which enables more variations of the attribute and hence more control options, e.g., radio buttons or sliders.

To the best of our knowledge, we are the first to incorporate a pairwise ranker in GAN for image generation task. In contrast, a recent work \cite{RankGAN} proposes a method, alias RankGAN, which substitutes the discriminator of GAN by a ranker for generating high-quality natural language descriptions, where the ranker's objective is to order human-written sentences higher than machine-written sentences with respect to reference sentences, and the generator's objective is to produce a synthetic sentence that receives higher ranking score than those drawn from the real data. If we were to project this method on image generation task, the ranker will be modelled to order real image higher than generated ones. But, in our work, the ranker will play a different role: it orders the images according to their present semantic attributes rather than the quality of images.

\paragraph{Image Editing:}
Image editing methods have a long history in the research community. Recently, CNN based approaches have shown promising results in image editing tasks such as image colorization \cite{Color1,Color2,Color3},  filtering \cite{Filter} and inpainting \cite{Inpainting}. These methods use an unsupervised training protocol during the reconstruction of the image, that may not capture important semantic contents. 

A handful works are interested in using deep generative models for image editing tasks. iGAN \cite{IGAN} allows the user to impose color and shape constraints on an input image, then reformulates them into an optimization problem to find the best latent code of GAN satisfying these constraints. The generated image motion and color flow are transferred during the interpolation to the real image. Similarly, the Neural photo editor \cite{IAN} proposes an interface for portrait editing using an hybrid VAE/GAN architecture. 

Aiming for high semantic level image editing, the Invertible Conditional GAN \cite{ICGAN} train separately a CGAN on the attributes of interest and an encoder that maps the input image to the latent space of CGAN so that it can be modified using the attribute latent variables. In the same context, CFGAN \cite{CFGAN} relies on iGAN's \cite{IGAN} approach to estimate the latent variables of an input image in order to edit its attributes, while the Fader network \cite{Fader} can directly manipulate the attributes of the input image since the encoder is incorporated in the training process.

\paragraph{Relative Attributes:} 
Visual concepts can be represented or described by semantic attributes. In early studies, binary attributes describing the presence of an attribute showed excellent performance in object recognition \cite{Recognition} and action recognition \cite{Action}. However, in order to quantify the strength of an attribute, we should aim for a better representation. D. Parikh~\etal \cite{Relative} use relative attributes to learn a global ranking function on images using constraints describing the relative emphasis of attributes (e.g. pairwise comparison of images). This approach is regarded as solving a learning-to-rank problem where a linear function is learned based on RankSVM \cite{RankSVM}. Similarly, RankNet \cite{RankNet} uses a neural network to model the ranking function using gradient descent methods. Also, Tompkin~\etal's Criteria Sliders \cite{Tompkin:2017:BMVC} applies semi-supervised learning to learn user-defined ranking functions. These relative attributes algorithms focus on predicting attributes on existing data entries. Our algorithm can be regarded as an extension enabling to continuously synthesize new data.  

%------------------------------------------------------------------------
\section{Approach}
In this section, we describe our contribution: ranking conditional GAN (RankCGAN), which allows image synthesis and image editing to be controlled by semantic attributes using subjectively specified scales. We first outline the generative CGAN \cite{CGAN}; then we outline the discriminative RankNet \cite{RankNet}; and last we show how to combine these distinct architectures to build RankCGAN.

\subsection{Generation by CGAN}
The CGAN \cite{CGAN} is an extension of GAN \cite{GAN} into a conditional setting. The generative adversarial network (GAN) is a generative model trained using a two-player minmax game. It  consists of two networks: a generator $G$ which outputs a generated image $G(z)$ given a latent variable $z \sim p_z(z)$, and a discriminator $D$ which is a binary classifier that outputs a probability $D(x)$ of the input image $x$ being real; that is, sampled from true data distribution $x \sim p_{data}(x)$. The minmax objective is formulated as the following:
\begin{equation}
    \min_{G}~\max_{D}\ \mathbb{E}_{x\sim p_{data}(x)}[log(D(x))]+\mathbb{E}_{z\sim p_z(z)}[log(1-D(G(z)))].
\end{equation}
In this model, there is no control over the latent space: the prior,  $p_z(.)$, on the multidimensional latent variable $z$ is often chosen as a multivariate normal $\mathcal{N}(0,I)$ or a multidimensional uniform $\mathcal{U}(-1,1)$ distribution.  Thus sampling generates images at random.

CGAN \cite{CGAN} affords some control over the generation process via the use of an additional latent variable, $r$, that is passed to both the generator and discriminator. The objective of CGAN can be expressed as:
\begin{equation}
    \min_{G}~\max_{D}\ \mathbb{E}_{r,x\sim p_{data}(r,x)}[log(D(r,x))]  + \mathbb{E}_{z\sim p_z(z), r\sim p_r(r)}[log(1-D(r,G(r,z)))].
\end{equation}

We can now consider the total latent space to be partitioned into two parts: a random subspace for $z$ which operates exactly as in standard GAN, and an attribute subspace containing $r$ over which the user has some control. The novel question we have answered in this paper is {\em how to provide subjective control over this subspace?}, our answer depends on an ability to discriminatively rank data.

\subsection{Discrimination by RankNet}

RankNet~\cite{RankNet} is a discriminative architecture in the sense that it classifies a pair of inputs, $x_i$  and $x_j$ according to their rank order: $x_i \triangleright x_j$ or $x_j \triangleright x_i$. As proposed in \cite{DeepRel}, the architecture comprises two parts:  (i) a mapping of any input $x$ to a feature vector $v(x)$; and (ii) a ranking function $R$ over these feature vectors. These components are modelled by convolutional layers. We use the shorthand $x_i \triangleright x_j$ or $R(x_i) > R(x_j)$ in place of $R(v(x_i)) > R(v(x_j))$, meaning that input $x_i$ is ranked higher than input $x_j$.

The convolutional layers in the first part can come pre-trained, or could be trained {\em in-situ}, either way the focus here is on setting the convolutional layer parameters defining the ranking function $R$. The training is supervised by a set of triplets $\{(x^{(1)}_i,x^{(2)}_i, y_i)\}_{i=1}^P$ where $P$ is the dataset size, $(x^{(1)}_i,x^{(2)}_i)$ pair of images and $y_i \in \{0,1\}$ a binary label indicating whether the image $x^{(1)}_i$ exhibits more of some attribute than $x^{(2)}_i$, or not. The loss function for a pair of images $(x^{(1)}_i,x^{(2)}_i)$ along with the target label $y_i$ is defined as a cross binary entropy:
\begin{equation}
\mathcal{L}_R(x^{(1)}_i,x^{(2)}_i,y_i) = -y_i \log(p_i) - (1-y_i) \log(1-p_i),
\end{equation}
where the posterior probabilities $p_i = P(x^{(1)}_i \triangleright x^{(2)}_i)$ make use of the estimated ranking scores
\begin{equation}
p_i = \mathrm{sig}(R(x^{(1)}_i)-R(x^{(2)}_i)) := \frac{1}{1 +e^{-(R(x^{(1)}_i)-R(x^{(2)}_i))}}.
\end{equation}
In testing, the ranker provides a ranking score for the input image which can be used to infer a global attribute ordering.

\subsection{Ranking Conditional Generative Adversarial Networks}
Recall our aim: to generate images of a particular object class, controlled by one or more subjective attributes.
With CGAN and RankNet available, we are now in a position to construct an architecture for this aim, an architecture that is a Ranking Conditional Generative Adversarial Network: which we call RankCGAN.
The key novelty of RankCGAN is the addition of an explicit ranking component to perform an end-to-end training with the generator and the discriminator. This training scheme is intended to put semantic ordering constraints in the generation process with respect to input latent variables.

\begin{figure*}[t]
\includegraphics[width=0.95\linewidth]{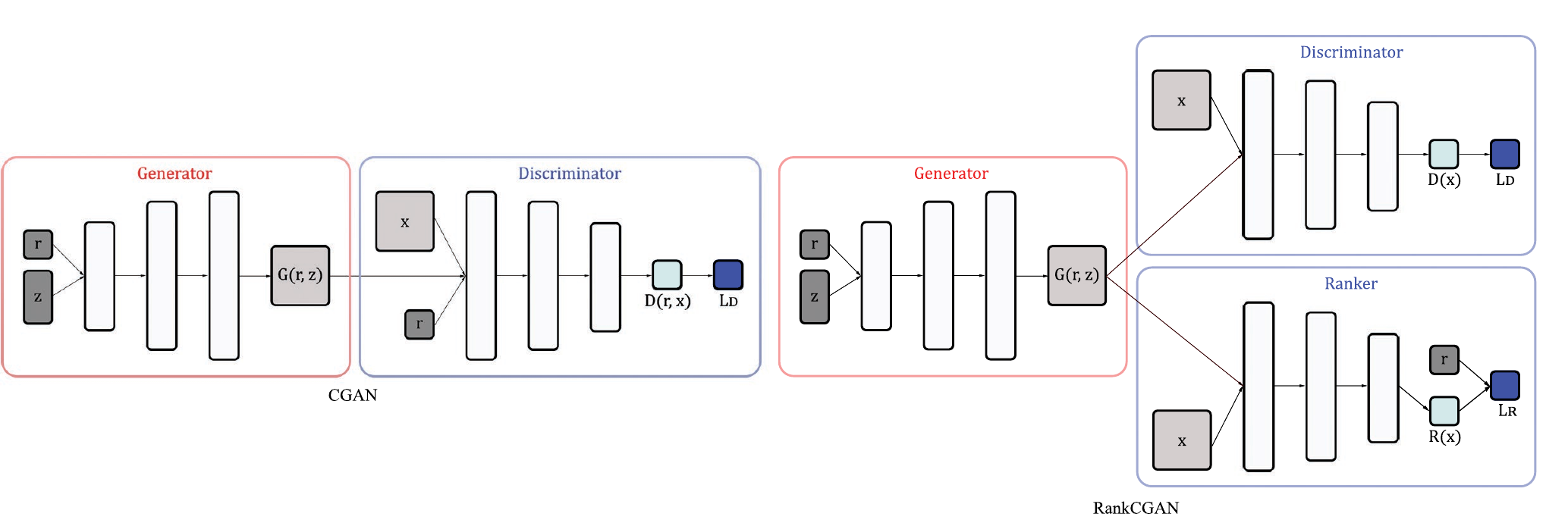}
\centering
%\caption{The difference between CGAN and RankCGAN architectures. The dark grey represents the latent variable, light grey represents the observable images, the scratched light grey indicates the output of the discriminator and the ranker, and the scratched dark grey indicates their loss functions.}
\caption{The difference between CGAN and RankCGAN architectures. The dark grey represents the latent variable, light grey represents the observable images, the light blue indicates the output of the discriminator and the ranker, and the dark blue indicates their loss functions.}
\label{fig:GAN}
\end{figure*}

A fundamental difference between RankCGAN and CGAN is in the use of the matching-aware discriminator \cite{Text2Image} in CGAN to define a real image as a one from the dataset with the correct label by incorporating the label in the discriminator. RankCGAN does not use this setting because it is trained on continuous controlling variables rather than binary labels.

As illustrated in Figure \ref{fig:GAN}, the architecture of RankCGAN is composed of three parts: the generator $G$, the discriminator $D$, and the ranker $R$. This architecture is versatile enough to rank over a line using one semantic attribute, specify points in plane using two semantic attributes, and in principle could operate in three or more dimensions. For simplicity's sake, we open our description with the one-dimensional case, followed by a generalization to $n$ dimensions. Finally we augment RankCGAN with an encoder that allows users to specify the subjective degree of semantic attributes, and in that way control image editing.

\subsubsection{The one-attribute case}

The generator $G$ takes two inputs, $z \sim \mathcal{N}(0,I)$ which is the unconditional latent vector, and $r \sim \mathcal{U}(-1,1)$ which is the latent variable controlling the attribute. The generator outputs an image $x = G(r,z)$. This image is input not just to a discriminating unit, $D(x)$, as in CGAN, but also to a ranking unit $R(x)$; as seen in Figure~\ref{fig:GAN}. Consequently, the architecture has three sets of parameters $\Theta_G$, $\Theta_D$, and $\Theta_R$ for the generator, discriminator, and ranker, respectively. The values of these parameters are determined by a training that involves three loss functions, one for each unit. 

The training is supervised in nature. As for RankNet,  let $\{(x^{(1)}_i,x^{(2)}_i, y_i)\}_{i=1}^P$ be pairwise comparisons and $\{x_i\}_{i=1}^N$ an image dataset of size $N$. These datasets are used in a mini-batch training scheme, of size $B$.
We defined three loss functions, one for each task: generation ($\mathcal{L}_G$), discrimination ($\mathcal{L}_D$), and ranking ($\mathcal{L}_R$).\\
For generation:
\begin{equation}
\mathcal{L}_G(I) = - \frac{1}{B} \sum_{i=1}^B log[D(I_i)], 
\end{equation}
such that ${\{I_i\}_{i=1}^{B}} = {\{G(r_i,z_i)\}_{i=1}^{B}}$. \\
For discrimination:
\begin{equation}
    \mathcal{L}_D(I) = -\frac{1}{B}\sum_{i=1}^B\left(t \log[D(I_i)] + (1-t) \log[1-D(I_i)]\right),\\
\end{equation}
where
$%\begin{equation}
t = \begin{cases}
1 & \text{if }  {\{I_i\}}_{i=1}^{B}={\{x_i\}}_{i=1}^{B},\\
0 & \text{if } {\{I_i\}}_{i=1}^{B}={\{G(r_i,z_i)\}}_{i=1}^{B}.
\end{cases}
$%\end{equation}
\\
For ranking:
\begin{equation}
\label{eq:7}
    \mathcal{L}_R(I^{(1)},I^{(2)},l) = -\frac{2}{B}\sum_{i=1}^{B/2}\left(l_i \log[\mathrm{sig}(R(I^{(1)}_i)-R(I^{(2)}_i))]
    + (1-l_i) \log[1-(\mathrm{sig}(R(I^{(1)}_i)-R(I^{(2)}_i)))]\right),
\end{equation}
with
\begin{align}
	{\left\{(I^{(1)}_i,I^{(2)}_i, l_i)\right\}}_{i=1}^{B/2}&=
	\begin{cases}
	\left\{(x^{(1)}_i,x^{(2)}_i, y_i)\right\}_{i=1}^{B/2},~\text{`for real images'};\\
	{\left\{(G(r_i^{(1)},z_i^{(1)}),G(r_i^{(2)},z_i^{(2)}),f(r_i^{(1)},r_i^{(2)}))\right\}}_{i=1}^{B/2},~\text{`for synthesised images'};
	\end{cases}
	\nonumber\\
	f(r_i^{(1)},r_i^{(2)})&=
\begin{cases}
1 & \text{if } r_i^{(1)}>r_i^{(2)},\\
0 & \text{else}.
\end{cases}
\end{align}
Based on these loss functions, the training algorithm for RankCGAN is defined by the adversarial training procedure presented as Algorithm~\ref{algo:RankCGAN} RankCGAN. The hyperparameter $\lambda$ controls the contribution of the ranker-discriminator during the updates of the generator.

\subsubsection{Multiple semantic attributes}

An interesting feature about RankCGAN is the ability to extend the model to multiple attributes. The architecture design and the training procedure remain intact, the only differences are the incorporation of new latent variables that control the additional attributes and the structure of the ranker which outputs a vector of ranking score with respect to each attribute. There are two different ways of designing a multi-attribute ranker, either using a separate ranking layer for each attribute, or a single ranking layer shared between all the attributes. Let $\{(x_i^{(1)}, x_i^{(2)}, \textbf{y}_i)\}_{i=1}^{P}$ be pairwise comparisons where $\textbf{y}_i$ is a vector of binary labels indicating whether $x_i^{(1)} \triangleright x_i^{(2)}$ or not with respect to all $S$ attributes. We define the loss function of ranking:
\begin{equation}
\mathcal{L}_R(I^{(1)}, I^{(2)}, l) = \sum_{j=1}^{S} \mathcal{L}_{R_j}(I^{(1)}, I^{(2)}, l_j),
\end{equation}  
where $ \mathcal{L}_{R_j}(I^{(1)}, I^{(2)}, l_j)$ is the ranking loss with respect to the attribute $j$, which is defined similarly to Equation \eqref{eq:7}:
 
\begin{align}
\mathcal{L}_{R_j}(I^{(1)},&I^{(2)},l_j) = -\frac{2}{B}\sum_{i=1}^{B/2}\left(l_{ij} \log[\mathrm{sig}(R(I^{(1)}_i)-R(I^{(2)}_i))]
    + (1-l_{ij}) \log[1-(\mathrm{sig}(R(I^{(1)}_i)-R(I^{(2)}_i)))]\right),
\end{align}

such that
\begin{align}
	{\left\{(I^{(1)}_i,I^{(2)}_i, l_{ij})\right\}}_{i=1}^{B/2}&=
	\begin{cases}
	{\left\{(x^{(1)}_i,x^{(2)}_i, y_{ij})\right\}}_{i=1}^{B/2},~\text{`for real images'};\\
	{\left\{(G(r_{ij}^{(1)},z_i^{(1)}), G(r_{ij}^{(2)},z_i^{(2)}),f(r_{ij}^{(1)},r_{ij}^{(2)}))\right\}}_{i=1}^{B/2},~\text{`for synthesised images'};
	\end{cases}
	\nonumber\\
f(r_{ij}^{(1)},r_{ij}^{(2)})&=
\begin{cases}
1 & \text{if } r_{ij}^{(1)}>r_{ij}^{(2)},\\
0 & \text{else}.
\end{cases}
\end{align}

with $y_{ij}~j$-th element in $\textbf{y}_i$ and $r_{ij}~i$-th input in the mini-batch, associated to the $j$-th attribute latent variable.
\begin{algorithm}
\caption{RankCGAN}
\label{algo:RankCGAN}
\begin{algorithmic}
\STATE \textbf{Set} the learning rate $\eta$, the batch size $B$, and the training iterations $S$
\STATE \textbf{Initialize} each network parameters $\Theta_D, \Theta_R, \Theta_G$
\STATE \textbf{Data} Images Set ${\{x_i\}}_{i=1}^{N}$, Pairs Set ${\left\{(x^{(1)}_i,x^{(2)}_i,y_i)\right\}}_{i=1}^{P}$
\FOR{n=1 \TO S}
\STATE Get real images mini-batches:\\ $x_{real} = {\{x_i\}}_{i=1}^{B}$,\\ $x_{real}^{(pair)} = {\left\{(x^{(1)}_i,x^{(2)}_i,y_i)\right\}}_{i=1}^{B/2}$ 
\STATE Get fake images mini-batches: \\$x_{fake} = {\{G(r_i,z_i)\}}_{i=1}^{B},$\\$x_{fake}^{(pair)} = {\left\{\left(G(r_i^{(1)},z_i^{(1)}),G(r_i^{(2)},z_i^{(2)}),f(r_i^{(1)},r_i^{(2)})\right)\right\}}_{i=1}^{B/2}$\\[0.1cm]
\STATE \textbf{Update the discriminator D:}\\
\STATE ~~~~~$\Theta_D \leftarrow \Theta_D - \eta \cdot \left( \frac{\partial L_D(x_{real})}{\partial \Theta_D} + \frac{\partial L_{D}(x_{fake})}{\partial \Theta_D}\right)$\\
\STATE \textbf{Update the ranker R:}\\
\STATE ~~~~~$\Theta_R \leftarrow \Theta_R - \eta \cdot  \frac{\partial L_R(x_{real}^{(pair)})}{\partial \Theta_R}$\\
\STATE \textbf{Update the generator G:}
\STATE  ~~~~~$\Theta_G \leftarrow \Theta_G - \eta \cdot \left( \frac{\partial L_G(x_{fake})}{\partial \Theta_G} + \lambda\cdot\frac{\partial L_R(x_{fake}^{(pair)})}{\partial \Theta_G} \right)$\\
\ENDFOR 
\end{algorithmic}
\end{algorithm}

\subsection{An Encoder for Image Editing}

Once trained, RankCGAN can be used for image synthesis. Since GAN lacks an inference mechanism, we use latent variables estimation for such tasks. Following previous works \cite{IGAN,ICGAN}, image editing means controlling some attributes of an image $x$ under a latent variable $r$ and random vector $z$, and generating the desired image by manipulating $r$. The approach consists of creating a dataset of size $M$ from the generated images and their latent variables $\{r_i,z_i,G(r_i,z_i)\}_{i=1}^{M}$ and training the encoders $E_r$, $E_z$ which encodes to $r$ and $z$ on this dataset. Their loss functions are defined in the mini-batch setting as follows:
\begin{equation}
\mathcal{L}_{Ez} = \frac{1}{B} \sum_{i=1}^{B} \parallel z_i - E_z(G(r_i,z_i)) \parallel_2^2.
\end{equation}
and
\begin{equation}
\mathcal{L}_{Er} = \frac{1}{B} \sum_{i=1}^{B} \parallel r_i - E_r(G(r_i,z_i)) \parallel_2^2.
\end{equation}
To reach a better estimation of $z$ and $r$ we can use the manifold projection method proposed in \cite{IGAN} which consists of optimizing the following function:
\begin{equation}
r^*,z^* = \underset{r,z}{\argmin} \parallel x - G(r,z) \parallel_2^2.   
\end{equation}
Unfortunately, this problem is non-convex, so that obtained estimates for $r^*$ and $z^*$ are strongly contingent upon the initial values of $r,z$; good initial values are provided by the encoders $E_r$, $E_z$. Therefore, we use manifold projection as a ``polishing'' step.

\section{Empirical Results} 
\label{sec:results}
In this section, we describe our experimental setup, quantitative and qualitative experiments, and outline applications in image generation, editing and semantic attribute transfer.

\begin{figure*}[t!]
\includegraphics[width=0.95\linewidth]{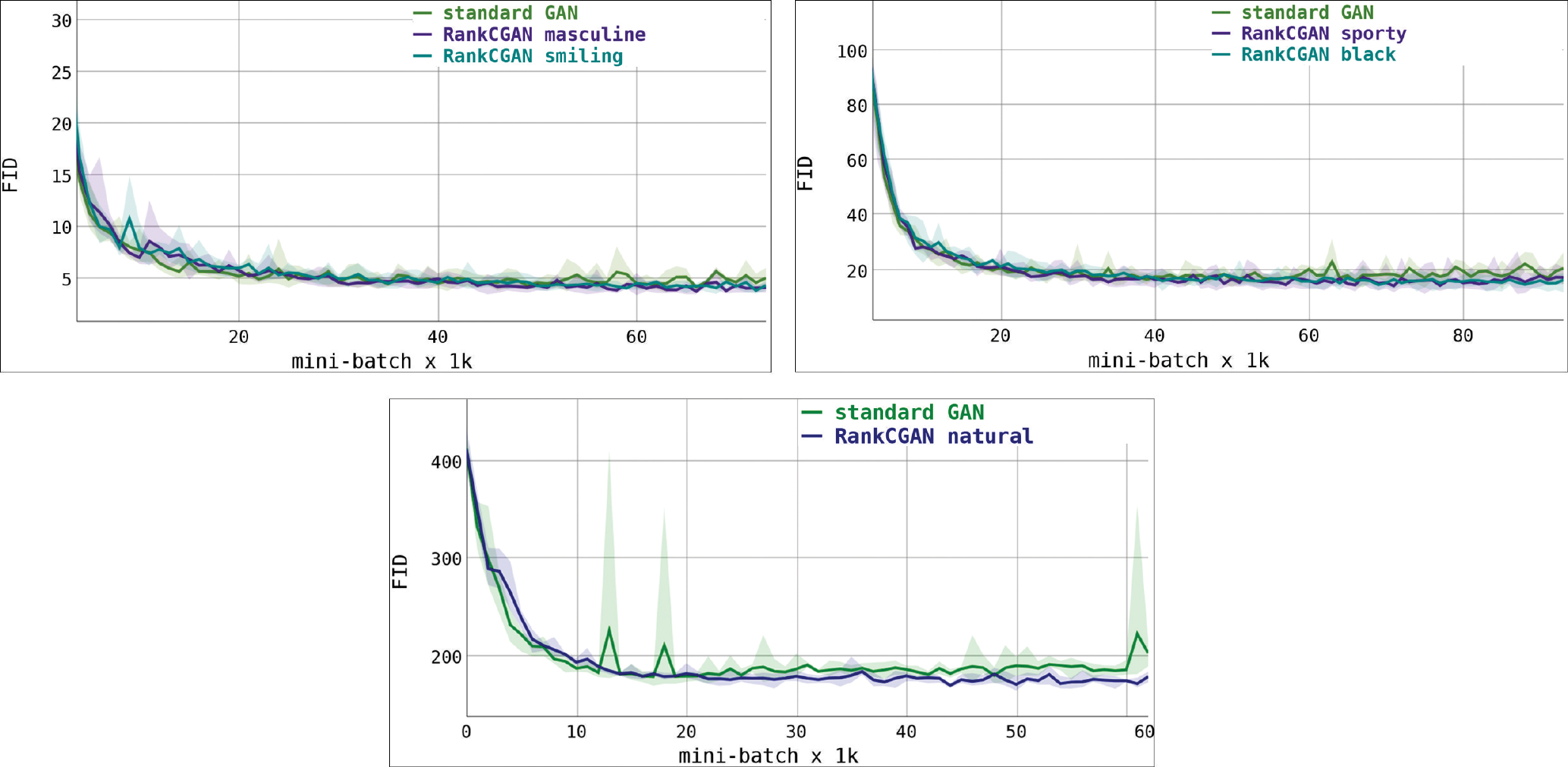}
\centering
\caption{Mean FID (solid line) surrounded by a shaded area bounded by the maximum and the minimum over 5 runs for GAN/RankCGAN on PubFig, OSR and UT-Zap50K datasets. \textbf{Top left:} PubFig with two RankCGANs trained on ``masculine'' and ``smiling'' attributes, starting at mini-batch update 3k for better visualisation. \textbf{Top Right:} UT-Zap50K with two RankCGANs trained on ``sporty'' and ``black'' attributes, starting at mini-batch update 3k for better visualisation. \textbf{Bottom:} OSR with one RankCGAN trained on ``natural'' attribute.}
\label{fig:quant}
\end{figure*}

\begin{figure*}[t]
\includegraphics[width=0.95\linewidth]{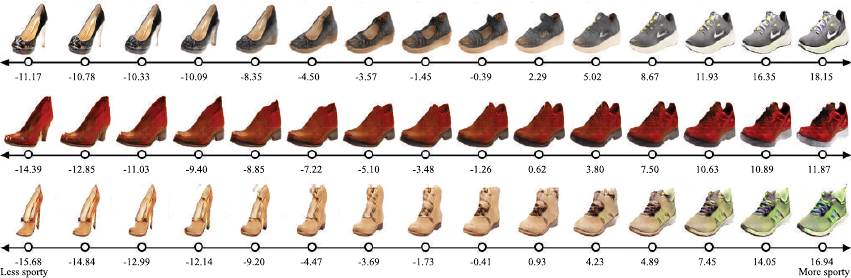}
\centering
\caption{Examples of generated shoe images with respect to ``sporty" attribute. The abscissa values represent the ranking score of each image}
\label{fig:shoe}
\end{figure*}

\subsection{Datasets} 

We used three datasets that provide relative attributes: the UT-Zap50K dataset \cite{Zappos}, the PubFig dataset \cite{Pubfig}, and the Outdoor Scene Recognition dataset (OSR) \cite{Relative}.

UT-Zap50K \cite{Zappos} consists of 50,025 shoe images from \emph{Zappos.com}. The shoes are in `catalog' style, being shown on white backgrounds, and are mostly of $150 \times 100$ pixels and have the same orientation. We rescaled them to $64 \times 64$ for GAN training purposes. The annotations for pairwise comparisons comprise 4 attributes (sporty, pointy, open, comfortable), with two collections: coarse and fine-grained pairs, where the coarse pairs are easy to visually discern than the fine-grained pairs. For this reason, we relied on the first collection in our experiments. It contains 1,500 to 1,800 ordered pairs per each attribute. Additionally, we defined the ``black'' attribute by comparing the color histogram of images.

We also conducted experiments on the PubFig dataset \cite{Pubfig} consisting of 15,738 face images, successfully downloaded, of different sizes rescaled to $64 \times 64$. A subset of PubFig dataset is used to build a relative attributes dataset \cite{Pubfig-p} containing 900 facial images of 60 categories and 29 attributes. The ordering of samples in this dataset is annotated in a category level, where all images in a specific category may be ranked higher, equal, or lower than all images in another category, with respect to an attribute. We used 50,000 pairwise image comparisons from the ordered categories, to train the ranker on a specific attribute.

Finally, we evaluated our approach on the Outdoor Scene Recognition dataset (OSR) \cite{Relative} consisting of 2,688 images from 8 scene categories and 6 attributes. Similarly to PubFig dataset, the attributes are defined in a category level, which enable to create 50,000 pairwise image comparisons to train the ranker on the desired attribute.

\subsection{Implementation}
Our RankCGAN is built on top of DCGAN \cite{DCGAN} implementation with the same hyperparameters setting and structure of the discriminator $D$ and the generator $G$. Besides, we modelled the encoders $E_r$, $E_z$, and the ranker $R$ with the same architecture of the discriminator $D$, except for the last sigmoid layer. We set the hyperparameter $\lambda=1$, learning rate to $0.0002$, mini-batch of size 64, and trained the networks using mini-batch stochastic gradient descent (SGD) with Adam optimizer \cite{Adam}. We trained the networks on UT-Zap50K and PubFig datasets for 300 epochs and on OSR dataset for 1,500 epochs.

\begin{figure*}[t]
\includegraphics[width=\linewidth]{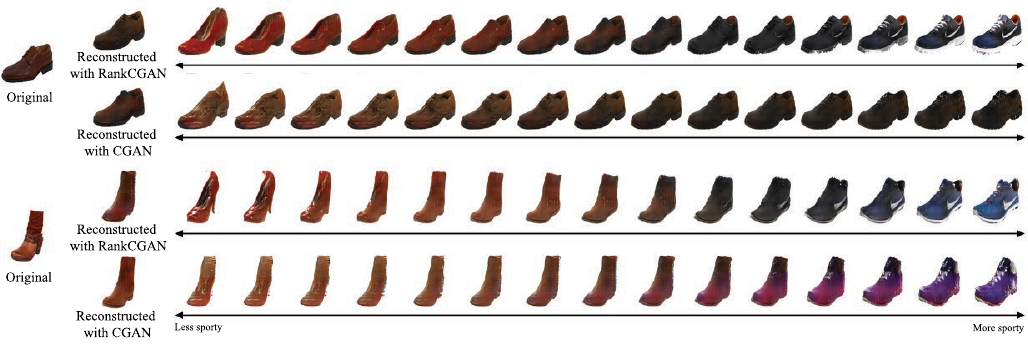}
\centering
\caption{Examples of the interpolation of similar images with respect to ``sporty" attribute using RankCGAN (rows 1, 3) and CGAN (rows 2, 4).}
\label{fig:comparison}
\end{figure*}

\begin{figure*}[t]
\includegraphics[width=0.95\linewidth]{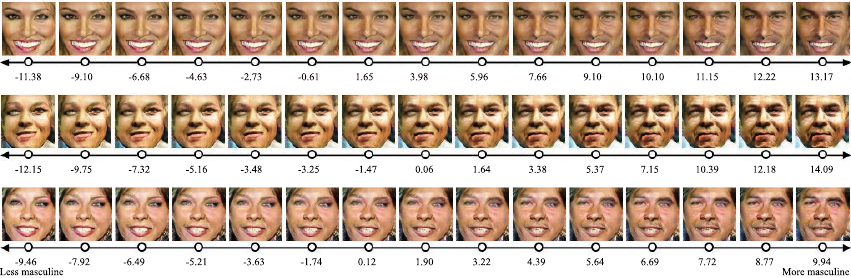}
\centering
\caption{Examples of generated face images with respect to ``masculine" attribute. The abscissa values represent the ranking score of each image.}
\label{fig:face}
\end{figure*}

\subsection{Quantitative Results}
\label{sec:quant}

A recent proposed evaluation method for GAN models is Fr\'echet Inception Distance (FID) \cite{FID}. In order to quantify the quality of generated images, they are embedded in a feature space given by the Inception Net pool 3 layer of 2048 dimension. Afterwards, the embedding layer is considered as a continuous multivariate Gaussian distribution, where the mean and covariance are estimated for both real and generated image. Then, the Fr\'echet Distance between these two Gaussians is used to quantify the quality of the samples.
In figure \ref{fig:quant}, we compare our proposed RankCGAN with the original GAN on all datasets to see whether the incorporation of the ranker has an impact on the quality of generated images. We calculate FID on every 1,000 mini-batch iterations with sampled generated images of the same size as the real images in the dataset. We notice that the standard GAN is a little faster at the beginning but eventually RankCGAN achieves slightly better or equal performance to GAN. The FID scores of RankCGAN and standard GAN are higher in OSR dataset due to the difficulty to generate its images while the FID is lower on PubFig dataset which shows that our model generative capability and quality are tied to the standard GAN.

\begin{figure*}[t]
\includegraphics[width=0.95\linewidth]{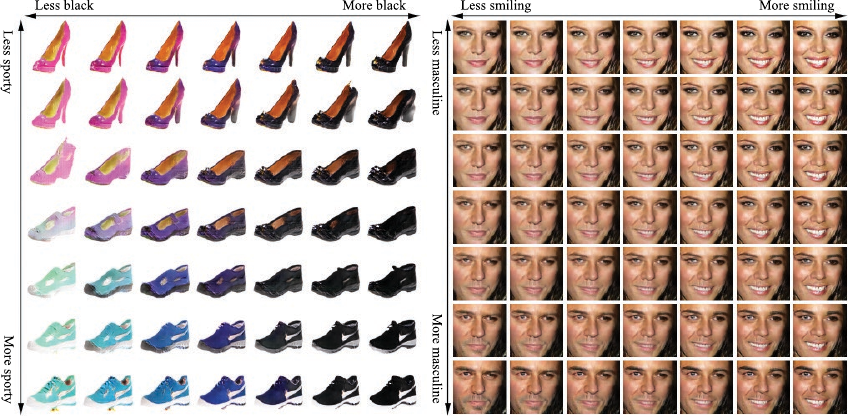}
\centering
\caption{Example of two-attributes interpolation on shoe (left) and face (right) images using (``sporty",``black") and (``masculine",``smiling") attributes.}
\label{fig:2D}
\end{figure*}

\subsection{Qualitative Results}
\label{sec:qual}

We would like an experiment that tests the following hypothesis: Incorporating an ranker brings an added value to the system which a CGAN is unable to achieve with its training procedure. Unfortunately, conducting an experiment to test the hypothesis is not straightforward. The problem is that RankCGAN requires ordering pairs for training some semantic attributes, whereas CGAN requires binary categorical labels that indicate its presence or absence.

To enable a comparison, we follow~\cite{Tompkin:2017:BMVC} method in mapping pairwise ordering to binary labels. We use a ranker with the same internal architecture as in RankCGAN, but independently trained on the ordered pairs to map each image in the dataset to a real ranking score. Then, all images with a negative score are said to not have the semantic property, while those with a positive score are said to possess it. The proportion of images with positive and negative scores in all the datasets is balanced around zero, which is used as a threshold. This method enables us to train CGAN on all datasets.

Qualitative results are shown in Figure~\ref{fig:comparison} for both RankCGAN and CGAN, in which the subjective scale runs from ``not sporty'' to ``sporty''. To use the trained networks to produce Figure~\ref{fig:comparison}, we used our encoders $E_r$, $E_z$ to estimate the latent variables $r$ and $z$ of a given real image. Then, we interpolated the image with respect to $r$ to each edge of the interval $[-1,1]$.

The results suggest that RankCGAN is capable of spanning a wider subjective interval than CGAN. Evidence for this is seen in the extreme ends of each interval. The top line (second row) of CGAN fails to reach shoes that can reasonably be called ``sporty'', while the bottom line (fourth row) fails to include a high-heel at the  ``not sporty''  extremity. In contrast, RankCGAN reaches a desirable shoe across the scale in both cases (first and third rows), we see a dress shoe at one end and a sporty shoe at the other. 

There is another, more subtle difference: the images produced by RankCGAN seem to change in a smoother way than those produced by CGAN. For example, the bottom line of CGAN shows many boots, with a change to sporty shoes coming late over the interval and occurring over a small subjective interval. This suggests that RankCGAN parametrizes the subjective space more uniformly than interpolating, as required when using CGAN.

\begin{figure*}[t]
\includegraphics[width=0.95\linewidth]{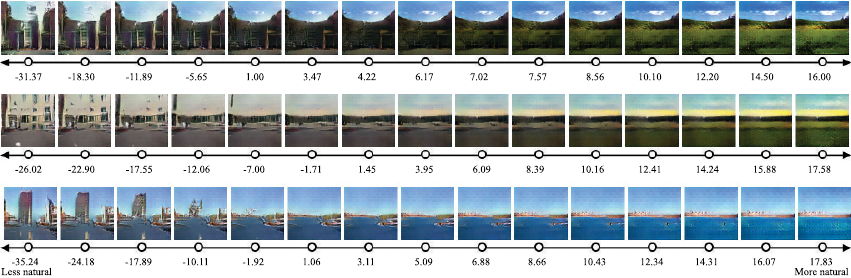}
\centering
\caption{Examples of generated scene images with respect to ``natural" attribute. The abscissa values represent the ranking score of each image.}
\label{fig:scene}
\end{figure*}
\begin{figure*}[t]
\includegraphics[width=0.8\linewidth]{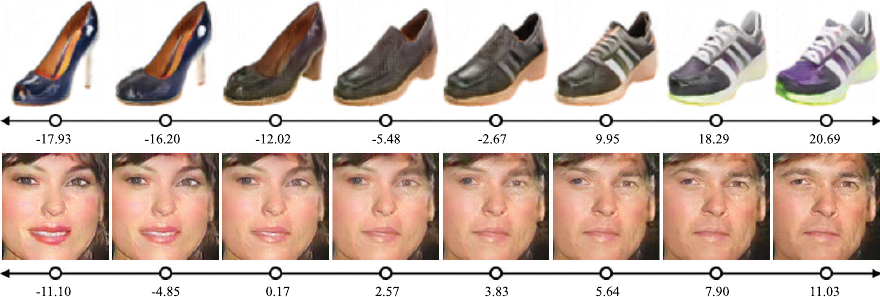}
\centering
\caption{Examples of $128 \times 128$ generated face and shoe images with respect to ``masculine'' and ``sporty'' attributes.}
\label{fig:high_res}
\end{figure*}
\begin{figure*}[t]
\includegraphics[width=0.95\linewidth]{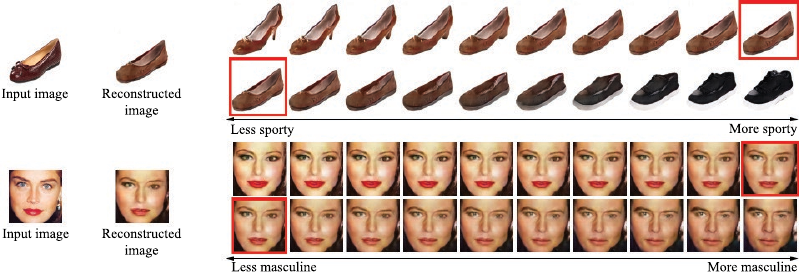}
\centering
\caption{Examples of image editing task, where the latent variables are estimated for an input image to perform the editing with respect to the ``sporty" attribute (top) and ``masculine" attribute (bottom).}
\label{fig:editing}
\end{figure*}

\subsection{Application: Image Generation}

Image generation is a basic expectation of any GAN. To  demonstrate this for RankCGAN, we choose single semantic attributes that span a line,  and pairs of semantic attributes that span a plane.  In all cases, the noise vector $z$ input to generator is held constant -- all variances in the generated images arise from changes in the semantic attributes alone.

Figures \ref{fig:shoe}, \ref{fig:face}, and \ref{fig:scene} show how the generated image varies with respect to the value of some subjective variable. These are ``sporty'', ``masculine'', and ``natural'', respectively; three examples are shown in each case. We note that plausible shoe and face images are generated at every point. Natural scenes are more difficult to generate for standard GAN, but our images are reasonable and progress from ``urban'' to ``natural'' in a pleasing manner. In all cases, the semantic attribute changes smoothly over subjective scale.

Additionally, Figure \ref{fig:high_res} demonstrates results on generating $128 \times 128$ images of faces and shoes datasets using the StackGAN method \cite{stackGAN}. The resolution is chosen due to our GPU memory limitation and the original size of images.

Concerning the strategy used for training RankCGAN on multiple attributes. We conducted experiments on both proposed strategies and noticed that we have similar results. The only reason behind choosing one over the other is the technical limitation. In fact, having a single shared layer between all attributes would require having pairwise comparisons labels with respect to every attribute for the same pair of images which is not always the existing case. For instance, in UT-Zap50K dataset, we have different pairs of image with respect to a specific semantic attribute. In this case the only resort is to train a separate ranking layer for each attribute. In Pubfig and OSR, the ordering is in category level, so we could use both training strategies. We just opted for the separate ranking layers strategy for all datasets.

Figure \ref{fig:2D} illustrates the image generation using two-attributes.
We used ``sporty", and ``black" for the shoe dataset, and ``masculine", and ``smiling" for face dataset. In both cases plausible images are generated, and the progression in both directions on both planes adheres to the subjective attributes in question. As with the one-dimensional cases, the semantic attribute changes smoothly in each direction. This is evidence that RankCGAN decorrelates multiple semantic attributes. 

\subsection{Application: Image Editing}

Image editing is an application that could be built into a system that supports user browsing. For example, a user might ask the system ``show me shoes that are less sporty than the one I see now''. The core of image editing is to map a given image onto the subjective scale by estimating its latent variables $r$, $z$ that produce a reconstructed image, and then move along the scale, one way or the other.
 
The image editing task was used to create Figure~\ref{fig:comparison} in the qualitative results (Section ~\ref{sec:qual}). In figure \ref{fig:editing}, we show image editing on shoes and on faces, using the ``sporty'' and ``masculine'' attributes respectively. The reconstructed  images are framed in red, and the images generated with subjectively less of the chosen attribute are shown on the top line, while the images generated to have subjectively more are on the bottom line. In both cases the noise vector $z$ was held constant, and only the semantic attribute $r$ varies across the learned subjective scale.

\subsection{Application: Semantic Attribute Transfer}

Our final application is semantic attribute transfer. The idea is to extract the subjective measure of a semantic attribute from one image, and apply that measure to another. This is possible because the encoder $E_r$ is able to quantify the semantic strength of an attribute in the image.  Indeed, this ability is used in all the results so far, generation excepted.

\begin{figure*}[h]
\includegraphics[width=0.6\linewidth]{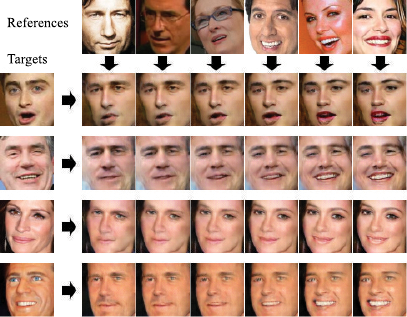}
\centering
\caption{Examples of ``smiling" attribute transfer task. The latent variable $r$ of reference images is estimated and then transferred to the target image in order to express the same strength of attribute.}
\label{fig:transfer}
\end{figure*}

In order to transfer an attribute, we quantify the conditional variable $r$ of the source image using the encoder $E_r$, then edit the target image with the new semantic value. Figure \ref{fig:transfer} shows some examples of this task. The reference images have been ordered, left to right, by increasing the subjective level of ``smiling'' . The corresponding semantic value is then used in conjunction with each of the target images to generate a new expression for the person in the picture.

We note that the target pictures do not have to be in a neutral expression to make this work. Indeed, some target images show people smiling, others do not. It is at best very difficult to see how to implement an application like this without a system that is able to model subjective scales, such as RankCGAN.

\section{Discussion and Conclusion}

We introduced RankCGAN, a novel GAN architecture that synthesises images with respect to semantic attributes defined relatively using a pairwise comparisons annotation. We showed through experiments that the design and training scheme of RankCGAN enable latent semantic variables to control the attribute strength in the generated images using a subjective scale.

Our proposed model is generic in the sense that it can be integrated into any extended CGAN model. It follows that, our model's generative power is tied to that of the GAN in terms of the quality of generated images, the diversity of the model and the evaluation methods. 

Possible extensions to this study consist of incorporating a filtering architecture, CFGAN \cite{CFGAN}, to enhance the RankCGAN controllability and also incorporating an encoder in the RankCGAN to perform an end-to-end training in order to improve image editing and attribute transfer tasks.

\section*{Acknowledgements}
We would like to thank James Tompkin for initial discussions about the main ideas of the paper. Yassir Saquil thanks the European Union's Horizon 2020 research and innovation programme under the Marie Sk\l{}odowska-Curie grant agreement No 665992 and the UK's EPSRC Centre for Doctoral Training in Digital Entertainment (CDE), EP/L016540/1. Kwang In Kim thanks EPSRC EP/M00533X/2 and RCUK EP/M023281/1.

\vfill
\bibliographystyle{ieee}
\bibliography{bib}
\end{document}